\documentclass[journal]{IEEEtran}

\usepackage{graphicx}
\usepackage{epsfig}
\usepackage{multirow}
\usepackage{amssymb}
\usepackage{amsmath}
\usepackage{amstext}
\usepackage{subcaption}
\usepackage{caption}
\usepackage{array}
\usepackage{tabulary}
\usepackage{makecell}
\renewcommand{\thetable}{\Roman{table}}

\ifCLASSINFOpdf

\else
 
\fi

\begin{document}

\title{Ensemble of Jointly Trained Deep Neural Network-Based Acoustic Models for Reverberant Speech Recognition}

\author{Jeehye~Lee, Myungin~Lee, and Joon-Hyuk~Chang,~\IEEEmembership{Senior Member,~IEEE}
\thanks{J. Lee, M. Lee, and J.-H. Chang are with Hanyang University, Seoul,
04763, Korea (e-mail : jchang@hanyang.ac.kr).}
}

\markboth{IEEE/ACM TRANSACTIONS ON AUDIO, SPEECH, AND LANGUAGE PROCESSING}%
{Lee and Chang: Ensemble of Jointly Trained Deep Neural Network-Based Acoustic Models for Reverberant Speech Recognition}

\maketitle

\begin{abstract}
Distant speech recognition is a challenge, particularly due to the corruption of speech signals by reverberation caused by large distances between the speaker and microphone. In order to cope with a wide range of reverberations in real-world situations, we present novel approaches for acoustic modeling including an ensemble of deep neural networks (DNNs) and an ensemble of jointly trained DNNs. First, multiple DNNs are established, each of which corresponds to a different reverberation time 60 ($\rm{RT_{60}}$) in a setup step. Also, each model in the ensemble of DNN acoustic models is further jointly trained, including both feature mapping and acoustic modeling, where the feature mapping is designed for the dereverberation as a front-end. In a testing phase, the two most likely DNNs are chosen from the DNN ensemble using maximum \textit{a posteriori} (MAP) probabilities, computed in an online fashion by using maximum likelihood (ML)-based blind $\rm{RT_{60}}$ estimation and then the posterior probability outputs from two DNNs are combined using the ML-based weights as a simple average. Extensive experiments demonstrate that the proposed approach leads to substantial improvements in speech recognition accuracy over the conventional DNN baseline systems under diverse reverberant conditions.
\end{abstract}


\IEEEpeerreviewmaketitle

\section{Introduction}
\IEEEPARstart{S}{ignals} originating from the same speech source usually appear differently due to a variety of acoustic reverberation effects. In speech recognition, these acoustic effects result in mismatches between trained speech recognition models and input speech. Two general approaches for reducing acoustic mismatch include feature mapping and model adaptation. In feature mapping techniques, input signal waveforms or feature vectors are converted to their enhanced versions before being fed into a recognizer. This is done using front-end processing techniques such as  packet loss concealment [1], linear filtering [2], [3], spectral enhancement [4]-[6], and feature enhancement [7], [8], which reflect long-term acoustic context. The first approach, termed linear filtering, dereverberates in time or transform domains, while spectral enhancement dereverberates corrupted power spectra. Additionally, feature enhancement aims to directly remove the effect of reverberation from corrupted feature vectors. Recently, deep neural networks (DNNs) have been used for matching reverberant speech to its anechoic version [9]-[14]. The mapper is trained with an architecture of multiple-layers, where the input is the spectral representation of reverberant speech and the desired output is that of anechoic speech. Hence, the feature mapping design can be treated as a problem during system identification, with a set of input and corresponding output feature vector sequences. Notice that this approach is still confined to the front-end of the speech recognition system, and thus the recognition results do not affect the design of the DNN-based feature enhancement. Another possible approach to reduce this mismatch is to train the acoustic model by using far more data under various reverberant conditions, which adjusts the speech recognition model parameters to more closely fit the input speech signal. This approach is termed the back-end-based speech recognition system, which includes a hidden Markov model (HMM) adaptation, such as the maximum likelihood linear regression (MLLR) [15] and the maximum \textit{a posteriori} (MAP) adaptation [16]. These techniques can also be used to mitigate the mismatch between clean HMMs and reverberant data, but their recognition performance in reverberant environments is often insufficient.

In recent years, acoustic modeling for the emission distribution of HMMs in speech recognition has been successfully replaced by neural networks with multiple-layers, which are generatively pre-trained without the use of discriminative information. Once generative pre-training is performed, discriminative fine-tuning (using back-propagation) adjusts the weights to improve their ability to predict a probability distribution over the states of monophone HMMs. More recently, neural network has been used to jointly train a single DNN for both feature mapping and acoustic modeling in noisy environments [14], [17], [18]. Instead of extracting acoustic features from enhanced speech, DNNs are employed as highly nonlinear mapping functions for the estimated clean speech features obtained from noisy speech. Then, a hybrid DNN architecture is designed in order to jointly train DNNs for both feature mapping and acoustic modeling. Consequently, the output layer of feature mapping becomes the input layer for acoustic modeling. Thus, joint training enables error back-propagation up to the feature mapping layer. To summarize the previous studies:

{\setlength{\parindent}{0cm}
1. Can we handle a wide range of sophisticated reverberation in real-world situations, using a single DNN-based acoustic model?
}

{\setlength{\parindent}{0cm}
2. No joint training techniques incorporating both feature mapping for dereverberation and acoustic modeling have been developed.
}

To answer the first question, it is worth mentioning the neural network ensemble, which builds a set of separately trained neural networks and combines the sets to form the unified prediction model [19]-[27]. Each trained neural network in the ensemble can serve a different role in modeling and can be applied to various deep learning systems to achieve greater recognition accuracy. Indeed, there have been efforts to build ensemble acoustic models (EAMs) by using neural networks for speech recognition. These techniques have generally exploited the sensitivity of acoustic models in order to yield inter-model diversity.

Motivated by insights mentioned above, we propose an ensemble of DNN acoustic models under a variety of reverberant conditions. Toward this end, we first split the reverberant data into multiple cases according to reverberation time 60 ($\rm{RT_{60}}$) in a training step. Next, one acoustic model is trained in a conventional manner for each $\rm{RT_{60}}$ range, where $\rm{RT_{60}}$ is one of the main parameters describing the reverberant degree. Once the EAMs using DNNs are trained, two types of DNNs are chosen in a probabilistic manner. This is done in a test step for which the probability is found in an independent module based on MAP criteria. As for the MAP criteria, maximum likelihood (ML) estimation is used for the blind estimation of $\rm{RT_{60}}$ [28], [29], which is based on a statistical model for representing sound decay under reverberant conditions. Indeed, the posterior probability outputs from two acoustic models are combined using a probabilistic average, and this combination yields superior performance when compared to a single acoustic model. In addition, inspired by the joint training technique [14], we propose jointly training each model in the ensemble for both feature mapping and acoustic modeling, which enables us to use the back-propagation algorithm up to the feature mapping layer. As a result, the feature mapping network is closely refined to the specified acoustic model in each network of ensemble models by connecting the output layer of the feature mapping DNN to the input layer of the DNN acoustic models. Each jointly trained DNN acoustic model is used again to develop the EAM, which is then selected in the same manner as the $\rm{RT_{60}}$-based MAP criteria.

The remainder of this paper is organized as follows. In the next section, we review the conventional methods for robust speech recognition. In Section III, we describe the design of the proposed ensemble joint acoustic model (EJAM) using ML and a model-selecting method. Extensive evaluation of the proposed method is discussed in Section IV, and conclusions are presented in Section V.

\section{REVIEW OF RELATED WORK}
We begin with a brief description of related work to provide sufficient background for understanding our approach. In the ensemble model, we first elaborate on the design of the ensemble classifier. In the second subsection, DNN joint training is introduced.
\subsection{Neural network ensemble}
\begin{figure*}[thb]
\begin{center}
\includegraphics[width=11cm]{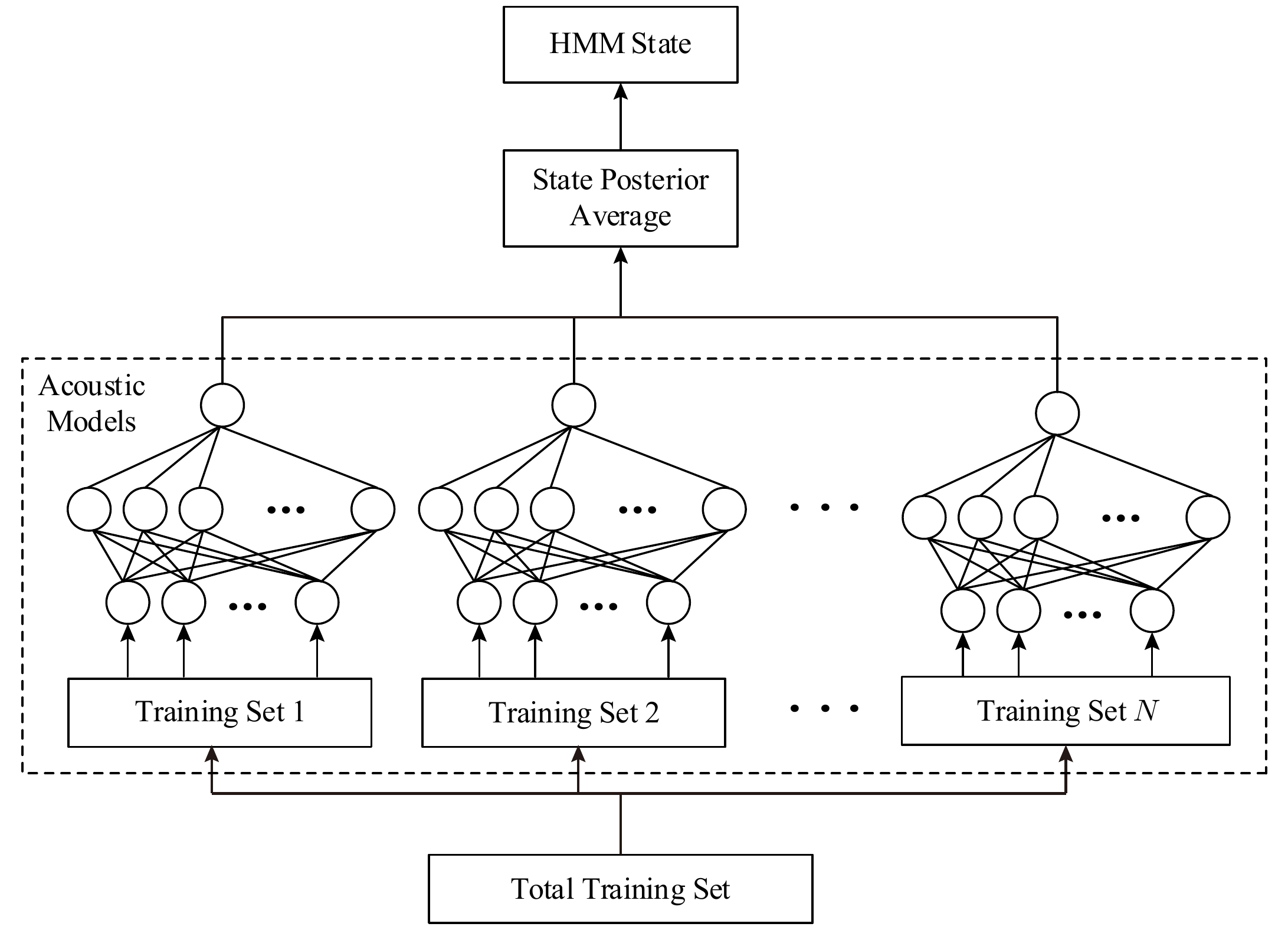}
\caption{Representation example of the ensemble acoustic model}
\end{center}
\end{figure*}
The ensemble classifier is known to be an effective approach for enhancing the recognition accuracy [21]-[23] for which individual models are independently designed, and the decoding word hypotheses of multiple models are combined to score the speech frames. In a recent study, multiple datasets were generated  through normalized noisy features by which beamforming and speech enhancement techniques are used, and additional speaker related features as well as other auxiliary features are also included [23]. One acoustic model is trained from each dataset, which builds up the acoustic model ensemble, as illustrated in Fig.\ 1. Then, the EAM outputs are combined via a simple posterior strategy, in which the output probability for the triphone HMM state $k$ is obtained as the average of the state posterior probabilities $p(k_n\mid \mathbf{x})$ of the $n$th acoustic model ($n=1,2,\ldots,N$) as given by:
\begin{eqnarray}
p(k_{\rm{EAM}}\mid \mathbf{x})=\frac{1}{N}\sum_{n=1}^{N}p(k_n\mid \mathbf{x})
\end{eqnarray}
where $\mathbf{x}$ is an input feature vector. The state posterior probabilities $p(k_{\rm{EAM}}\mid \mathbf{x})$ are utilized in decoding searches. In averaging posterior probabilities, simple averaging methods [21], [23] and weighted averaging method [22] have been presented. It is worth noting that the recognition performance of the combined model is enhanced when models are used from a variety of different sources. For a more detailed discussion of the EAM, please refer to [23].

\subsection{Joint training of DNNs}
\begin{figure}[thb]
\begin{center}
\includegraphics[width=9cm]{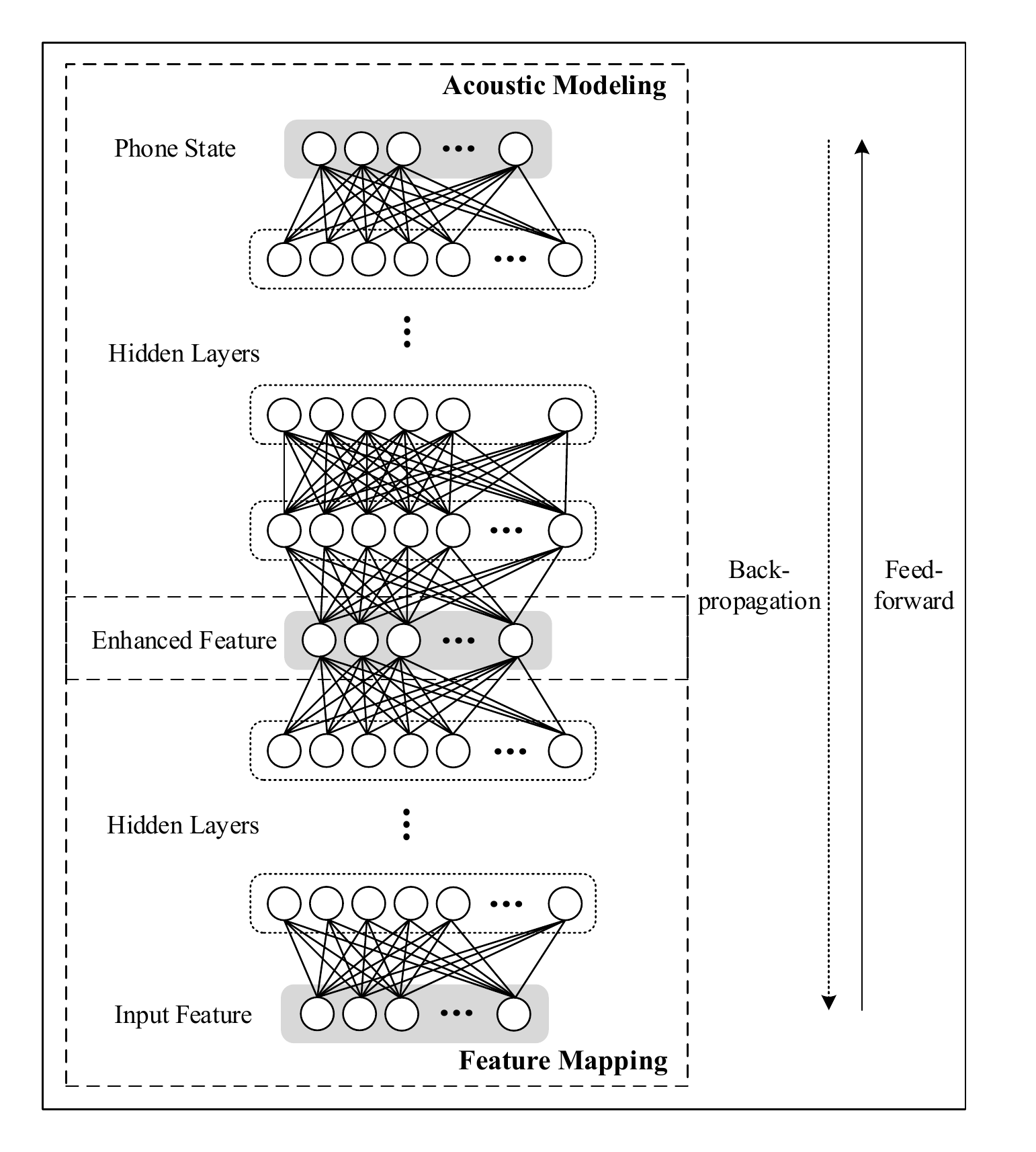}
\end{center}
\caption{Structure of the joint training method}
\end{figure}
Instead of extracting acoustic features from the speech waveform, the DNN is used as a nonlinear feature mapping function. Then, the joint training method is developed by integrating the DNN for feature mapping and the DNN for acoustic modeling, and subsequently jointly training the integrated DNN, as shown in Fig.\ 2. In feature mapping, the DNN is designed to estimate the clean speech feature from a noisy feature, for which minimizing the mean squared error (MSE) between the DNN output and reference target clean speech feature can be described as follows:
\begin{eqnarray}
E=\frac{1}{N}\sum_{n=1}^{N}\|{\hat{\mathbf{x}}_{n-\tau}^{n+\tau}}(\mathbf{y}_{n-\tau}^{n+\tau},\mathbf{W},\mathbf{b})-\mathbf{x}_{n-\tau}^{n+\tau}\|^2_2+\kappa\|\mathbf{W}\|^2_2
\end{eqnarray}
where $\hat{\mathbf{x}}_{n-\tau}^{n+\tau}$ and $\mathbf{x}_{n-\tau}^{n+\tau}$ are the $D(2\tau+1)$-dimensional vectors of the estimated and reference clean features for the $n$th frame, respectively. $\mathbf{y}_{n-\tau}^{n+\tau}$ is a $D(2\tau+1)$-dimensional vectors of input noisy features with the neighboring left and right $\tau$ frames as the acoustic context. Additionally, $\mathbf{W}$ and $\mathbf{b}$ denote the weight and bias parameters, respectively, with $\kappa$ representing the regularization weighting coefficient and $N$ denoting the mini-batch frame size. The acoustic model is then trained by using the enhanced features obtained from feature mapping and acoustic modeling DNN layers are stacked on top of the feature mapping layer. Naturally, the output layer of feature mapping becomes the input layer for acoustic modeling. Then, the error back-propagation algorithm is utilized in order to jointly train the single DNN, which includes both feature mapping and acoustic modeling. Consequently, the jointly trained DNN is advantageous, in that feature mapping is refined to acoustic modeling (and vice versa). This seamless connection between two DNNs permits high levels of recognition accuracy.

\section{PROPOSED ENSEMBLE JOINT ACOUSTIC MODEL FOR REVERBERANT SPEECH RECOGNITION}
In this section, we introduce the proposed algorithm for the DNN ensemble model and ensemble of jointly trained DNN models. We then present the manner in which to combine the acoustic model ensemble, for which the blind estimation of $\rm{RT_{60}}$ is described.

\subsection{Ensemble of DNN models}
\begin{figure*}[thb]
\begin{center}
\includegraphics[width=14cm]{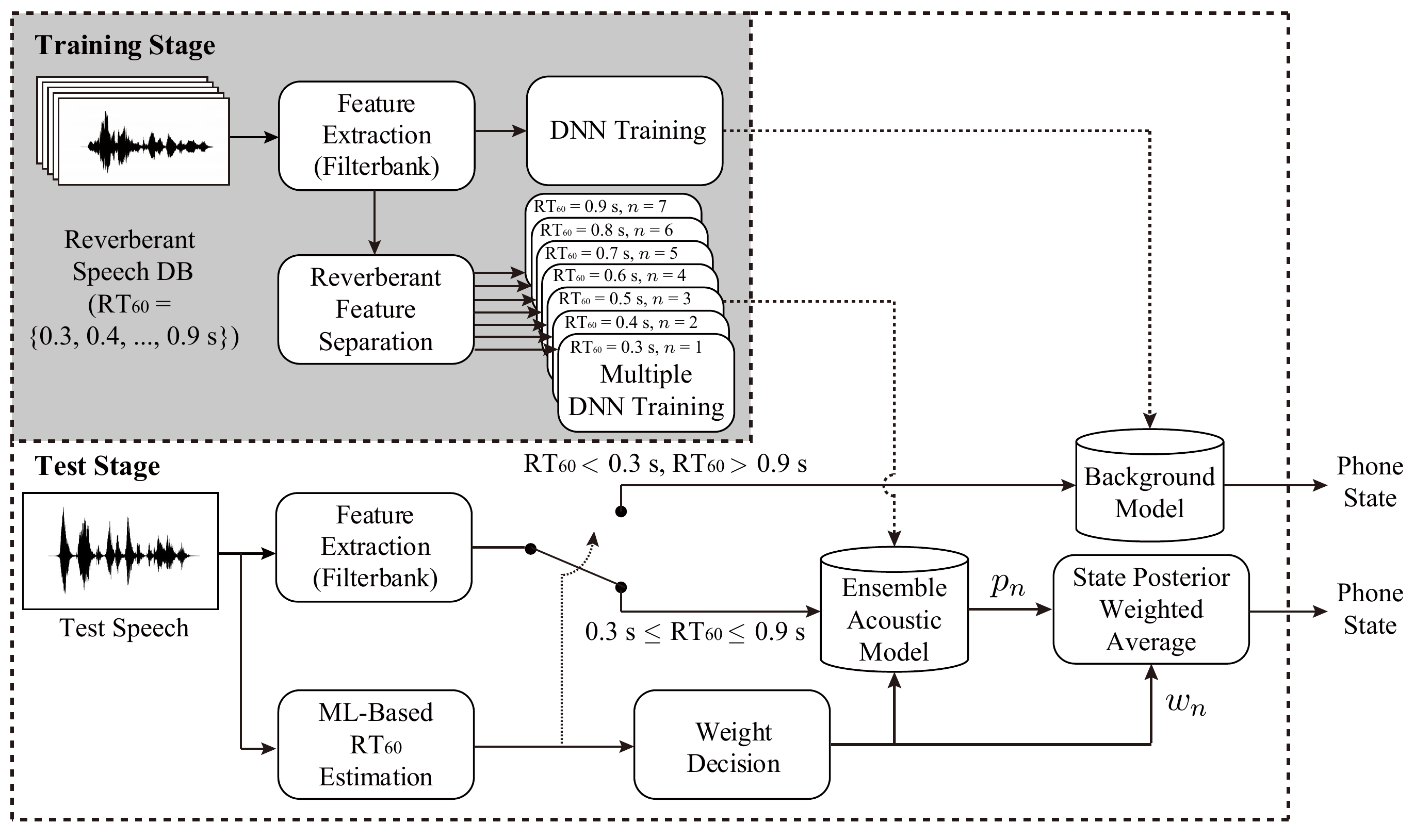}
\caption{Block diagram of the ensemble model}
\end{center}
\end{figure*}
In short, we design seven different classes of neural networks for acoustic modeling, each of which quantifies a unique reverberant condition. After filterbank feature extraction from reverberant speech, features are divided into multiple training sets. For this, the reverberant condition is divided into seven points, from $\rm{RT_{60}}=0.3\ s$ to $\rm{RT_{60}}=0.9\ s$, in increments of $\rm0.1\ s$. From the dataset for each point, the DNN-based EAM is separately established using the pre-training and fine-tuning techniques as described in the previous section. As a result, the EAM consists of seven different DNNs, as shown in Fig.\ 3. In a test step, the posterior weighted averaging technique is then applied to combine the results of the DNN ensemble. If we assume that $N$ acoustic models are generated from $N$ multiple datasets, the output probability at the $n$th acoustic model is defined by $p_n=p(k_n\mid \mathbf{x})$, where $k$ denotes the HMM states. For the EAM, the final posterior probability is computed with $m_1, m_2\in\{n\mid 1,2,\ldots,N\}$ as
\begin{eqnarray}
P(k_{\rm{EAM}}\mid \mathbf{x})=w_{m_1}p(k_{m_1}\mid \mathbf{x})+w_{m_2}p(k_{m_2}\mid \mathbf{x}),
\end{eqnarray}
by fusing the two weighted outputs from the two most likely DNNs among seven different DNNs ensemble. The weights are determined by the MAP probability computation, which is given by the blind ML estimation of $\rm{RT_{60}}$ [29].

Specifically, to determine weights, we first need to blindly estimate $\rm{RT_{60}}$, a core parameter for characterizing reverberant environments. Blind estimation, which implies estimation of $\rm{RT_{60}}$, is performed without prior knowledge of speech sources or room geometry. Indeed, the diffuse tail of reverberation is instead mathematically modeled as a simplified noise decay curve [29]:
\begin{eqnarray}
d(k)=A_rv(k)e^{-\rho kT_s}\epsilon(k)
\end{eqnarray}
where $A_r$, $\rho$, and $\epsilon(k)$ are the real amplitude, decay rate, and unit step sequence, respectively. Additionally, $T_s$ denotes the sampling period, and $v(k)$ is a sequence of random variables. Since the sequence $d(k)$ for $k\in\{0,1,\ldots,N-1\}$ is modeled by $N$ independent random variables, we can find an ML estimator of $\rm{RT_{60}}$. Thus, the log-likelihood function can be expressed as:
\begin{multline}
\mathcal{L}(\rho)= \\
-\frac{N}{2}\Bigg((N-1)\ln(a)+\ln\Big(\frac{2\pi}{N}\sum_{i=1}^{N-1}a^{-2i}d^2(i)\Big)+1\Bigg)
\end{multline}
where $a=e^{-T_s\rho}$. Then, the decay rate $\rho$ is estimated based on the ML, as given by:
\begin{eqnarray}
\hat{\rho}^{\rm{(ML)}}=\max_{\rho}\{\mathcal{L}(\rho)\}
\end{eqnarray}
Then, $\hat{\rho}^{\rm{(ML)}}$ is transformed into $\rm{RT_{60}}$ as follows:
\begin{eqnarray}
\rm{RT_{60}}=\frac{3}{\rho\log_{10}e}\approx\frac{6.908}{\rho}.
\end{eqnarray}
Note that we use a downsampling operation and simple pre-selection of possible sound decays in order to reduce the computational complexity and increase the online estimation speed. Later, if the estimated $\rm{RT_{60}}$ falls to one of the seven points between $\rm0.3\ s$ and $\rm0.9\ s$, the two most likely DNNs ($m_1$ and $m_2$) are chosen as the candidates to be combined as in (3). Other DNN models are ignored to avoid the unwanted effect of outliers. For this, two weights are determined by the ratio of likelihoods, both computed as in (5) at the two most likely $\rm{RT_{60}}$ points, given by:
\begin{eqnarray}
w_{m_1}=\frac{\mathcal{L}(\rho_{m_1})}{\mathcal{L}(\rho_{m_1})+\mathcal{L}(\rho_{m_2})}
\end{eqnarray}
and
\begin{eqnarray}
w_{m_2}=\frac{\mathcal{L}(\rho_{m_2})}{\mathcal{L}(\rho_{m_1})+\mathcal{L}(\rho_{m_2})}.
\end{eqnarray}

\subsection{Ensemble of jointly trained DNN models}
\begin{figure*}[thb]
\begin{center}
\begin{subfigure}[b]{0.25\textwidth}
\includegraphics[width=\textwidth]{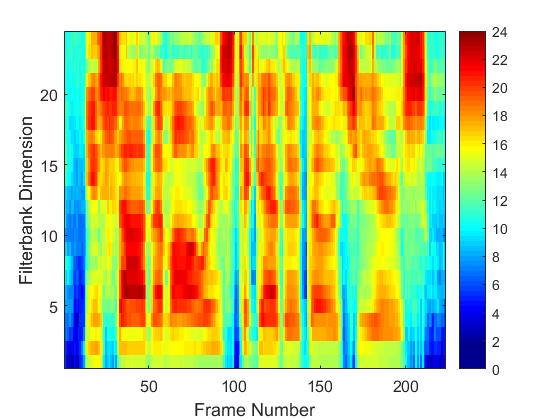}
\caption{Clean feature vectors}
\end{subfigure}

\begin{subfigure}[b]{0.23\textwidth}
\includegraphics[width=\textwidth]{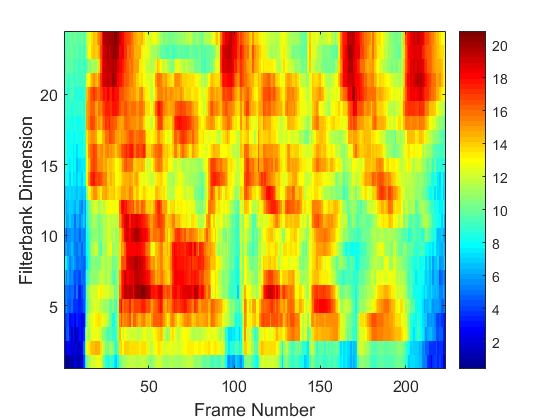}
\caption{\centering Reverberant \linebreak feature vectors ($\rm{RT_{60}}=0.3\ s$)}
\end{subfigure}
\begin{subfigure}[b]{0.23\textwidth}
\includegraphics[width=\textwidth]{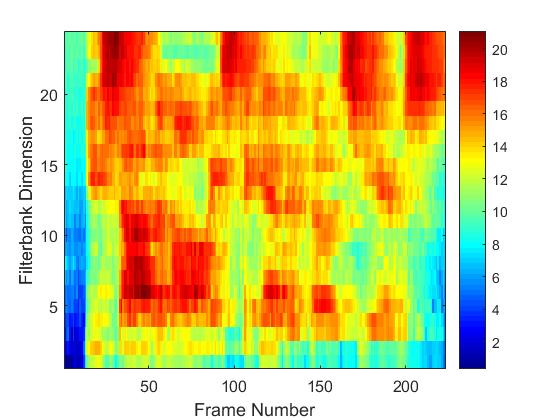}
\caption{\centering Reverberant \linebreak feature vectors ($\rm{RT_{60}}=0.5\ s$)}
\end{subfigure}
\begin{subfigure}[b]{0.23\textwidth}
\includegraphics[width=\textwidth]{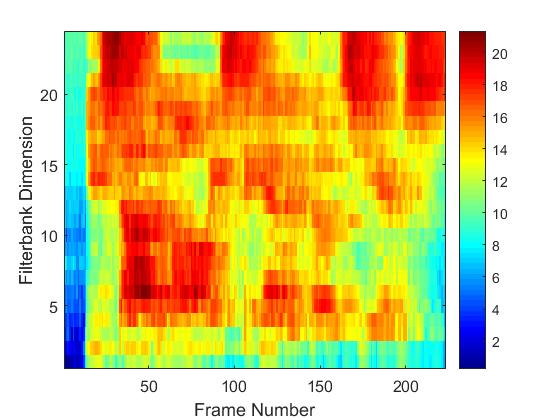}
\caption{\centering Reverberant \linebreak feature vectors ($\rm{RT_{60}}=0.7\ s$)}
\end{subfigure}
\begin{subfigure}[b]{0.23\textwidth}
\includegraphics[width=\textwidth]{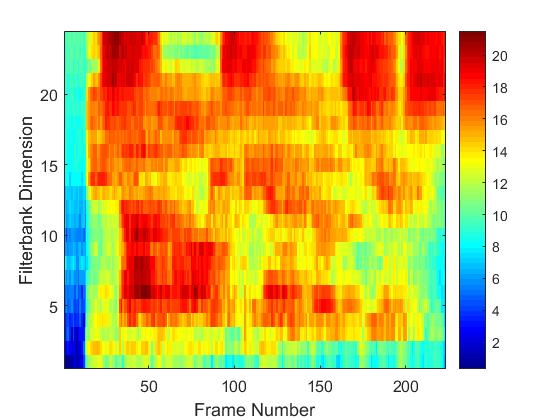}
\caption{\centering Reverberant \linebreak feature vectors ($\rm{RT_{60}}=0.9\ s$)}
\end{subfigure}
\begin{subfigure}[b]{0.23\textwidth}
\includegraphics[width=\textwidth]{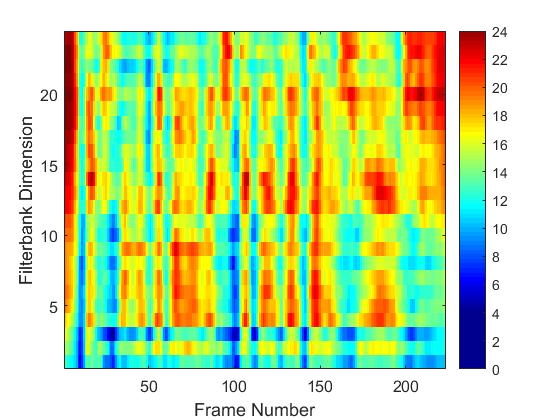}
\caption{\centering Dereverberated \linebreak feature vectors ($\rm{RT_{60}}=0.3\ s$)}
\end{subfigure}
\begin{subfigure}[b]{0.23\textwidth}
\includegraphics[width=\textwidth]{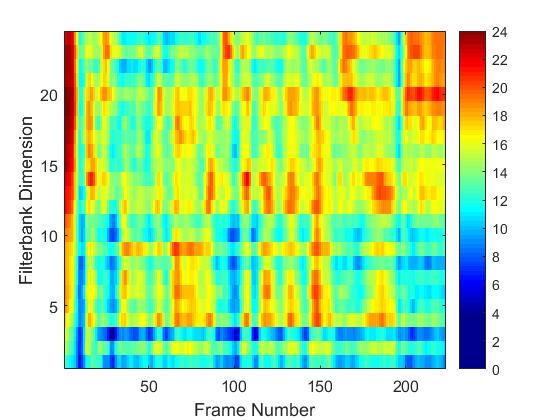}
\caption{\centering Dereverberated \linebreak feature vectors ($\rm{RT_{60}}=0.5\ s$)}
\end{subfigure}
\begin{subfigure}[b]{0.23\textwidth}
\includegraphics[width=\textwidth]{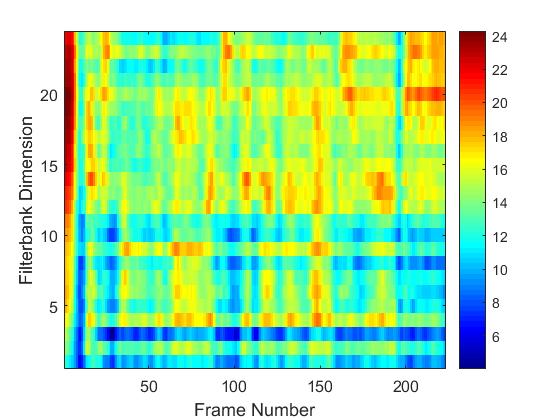}
\caption{\centering Dereverberated \linebreak feature vectors ($\rm{RT_{60}}=0.7\ s$)}
\end{subfigure}
\begin{subfigure}[b]{0.23\textwidth}
\includegraphics[width=\textwidth]{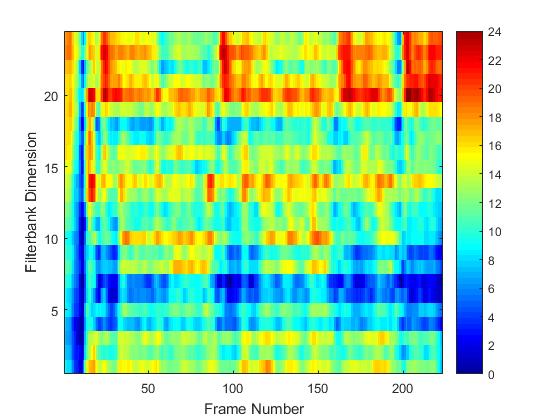}
\caption{\centering Dereverberated \linebreak feature vectors ($\rm{RT_{60}}=0.9\ s$)}
\end{subfigure}
\caption{An example of the feature mapping input and output layer}
\end{center}
\end{figure*}

In this subsection, we present the jointly trained DNN for the ensemble model by applying both feature mapping and acoustic modeling. Thus, the EJAM in this approach deals with reverberant speech at seven $\rm{RT_{60}}$ points, as in the previous EAM, which can be described as a convolution of clean speech with the room impulse response (RIR) at each $\rm{RT_{60}}$ point. The reverberant speech is employed as an input for feature mapping, for which the reverberant filterbank features are converted to the target clean filterbank features. Thus, design of the feature mapping rule is regarded as a problem of system identification, with a set of input and corresponding output feature vector sequences. Estimation of DNN model parameters relevant to feature mapping is performed based on three hidden layers with 2048 units in each layer by training to minimize the MSE function between the input and output. A total of seven DNN models are established from $\rm{RT_{60}}=0.3\ s$ to $\rm{RT_{60}}=0.9\ s$ in increments of $\rm0.1\ s$, as in the previous section. Since feature mapping can solve the problem of system identification, an example of feature mapping is shown in Fig.\ 4, which includes the filterbank features of clean speech and reverberant speech with $\rm{RT_{60}}=0.3\ s$, $\rm0.5\ s$, $\rm0.7\ s$, and $\rm0.9\ s$. Note that the prominent distinction between dereverberant speech and reverberant speech can be seen in the smearing of word boundaries.

\begin{figure*}[thb]
\begin{center}
\includegraphics[width=14cm]{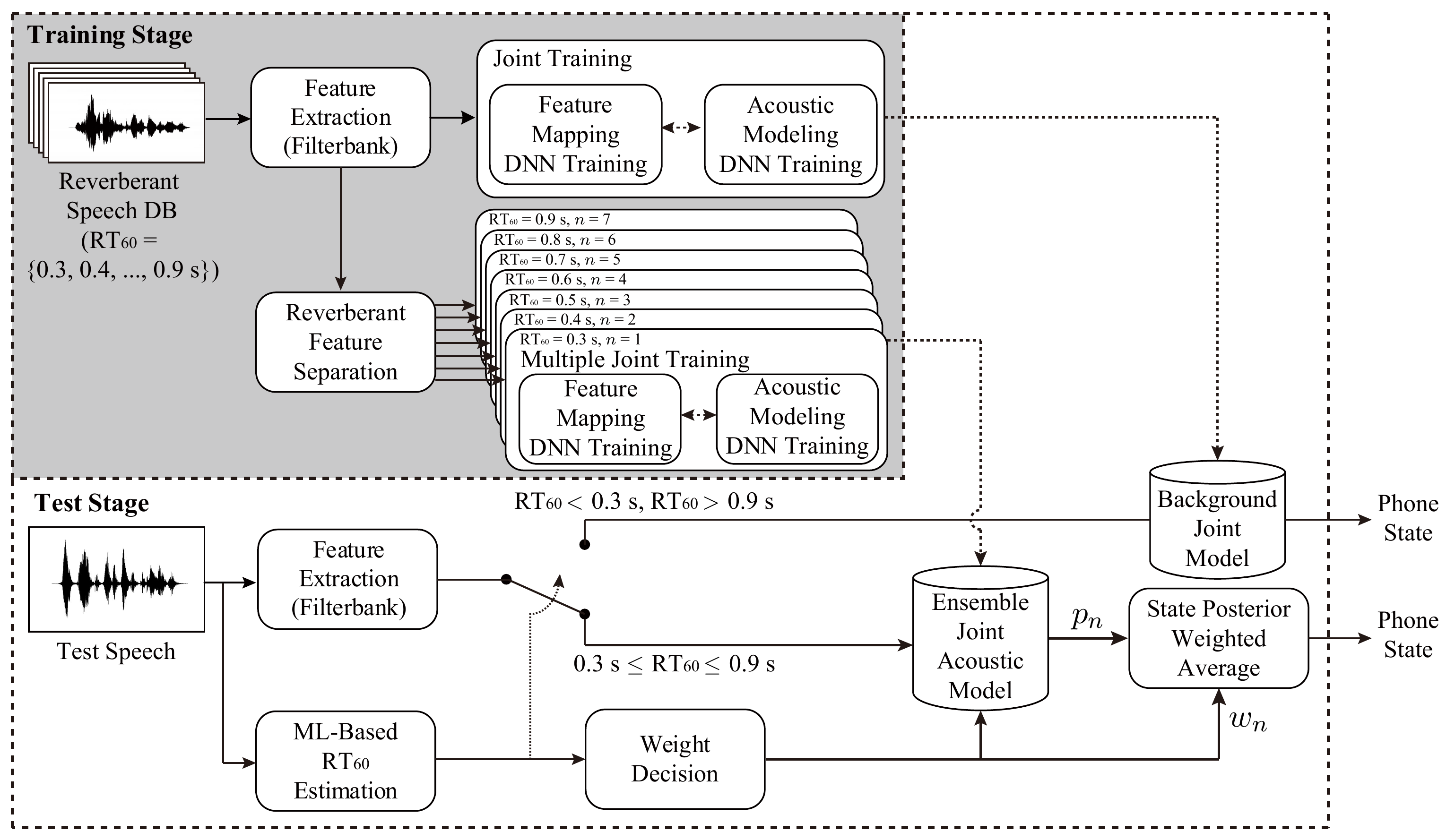}
\caption{Block diagram of the ensemble joint acoustic model}
\end{center}
\end{figure*}

Once the training for feature mapping is completed, the acoustic modeling layer at each $\rm{RT_{60}}$ point is independently trained using the output of the corresponding feature mapping layer. As shown in Fig.\ 2, the acoustic modeling DNN, which includes seven hidden layers with 2048 hidden units in each layer, is directly stacked on top of the feature mapping DNN. The combined ensemble DNN is jointly trained to minimize the total cross entropy error function by using the error back-propagation algorithm, as shown in Fig.\ 5. What remains is how to select the DNN from the EJAM, established on each $\rm{RT_{60}}$ point. Again, after the integration step for joint training is completed, the two joint DNNs that maximize the likelihood at each time are identified. Similar to the previous ensemble model, the ML-based weights are employed to achieve the final acoustic score $P(k_{\rm{EJAM}}\mid \mathbf{x})$ for each feature $\mathbf{x}$, as in (3).

\section{EXPERIMENTS}
This section describes the performance evaluation of the proposed EAM and EJAM for distant speech recognition. In order to assess the proposed method, we present a number of experiments performed in various reverberant environments. Moreover, the proposed method was compared with a conventional single DNN acoustic model and a single DNN acoustic model, which set the topological complexity equivalent to that of EAM [30].

\subsection{Experimental Setup}
\begin{figure}[tb]
\begin{center}
\includegraphics[height=5cm]{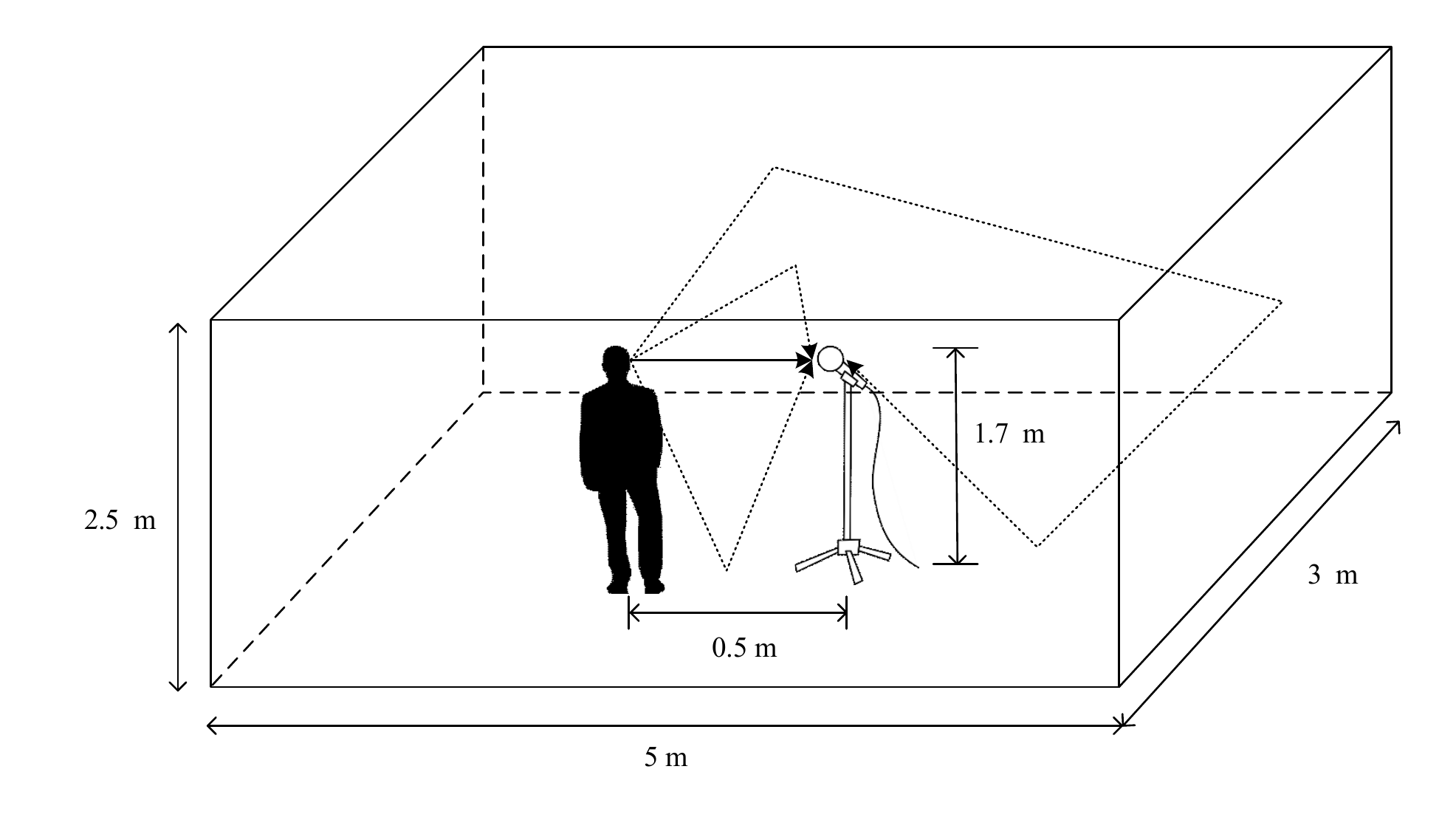}
\caption{Reverberant room environment}
\end{center}
\end{figure}
The proposed method was evaluated on TIMIT DB [31], which was first applied to the RIR generator [32] in order to simulate reverberant environments. As shown in Fig.\ 6, we generated a simulated room with size $5\times 3\times 2.5\ \rm{m^3}$, and varied reflection coefficients to yield specific $\rm{RT_{60}}$ conditions. The distance between the microphone and source was set to 50 cm in order to avoid close-talking scenarios. In order to simulate the system, reverberation times were increased from $\rm0.3\ s$ to $\rm0.9\ s$, which correspond to specific $\rm{RT_{60}}$ conditions. In time, 3696 utterances from TIMIT were employed to form the training set in each $\rm{RT_{60}}$ condition. As a result, the number of utterances used for training was $\rm{3320\times 7=23240}$, a number also used in building background models. The development set included 376 utterances mixed with seven $\rm{RT_{60}s}$ for cross validation. It is worth noting that the test set was prepared to include 192 reverberant sentences, corresponding to seventy $\rm{RT_{60}s}$ ($\rm0.3$-$\rm0.9\ s$, increment of $\rm0.01\ s$). Note that no overlap was permitted in utterances in training, development, or test sets. In the case of speech features, 72 components were generated, including 24-dimensional filterbank features and their first-order and second-order derivatives. Then, feature analysis was performed from each frame of 25 ms, with a 10 ms frame shift for the 16 kHz speech waveforms. Triphone modeling was used for HMMs, each of which was modeled by three emitting states. The number of tied states was 2021. We used a phone-based bigram language model estimated from training utterances. Training of HMM parameters and decoding for speech recognition was carried out using the Kaldi software [33].

\subsection{Speech recognition results of DNN-based ensemble models}
\begin{table*}[tbh]
\begin{center}
\caption{Performance [PER (\%)] comparison of the single background models and ensemble models in the matched test conditions}
\begin{tabular}{|c||c|c|c|c|c|c|c|c|c|c|c|c|c|c|c|c|} \hline
Model          & SBM & eSBM & \multicolumn{7}{c|}{EAM}   & \multicolumn{7}{c|}{EJAM} \\ \hline
$\rm{RT_{60}}$ &  \makecell{$\rm0.3\ s$\\-\\$\rm0.9\ s$}      & \makecell{$\rm0.3\ s$\\-\\$\rm0.9\ s$}              & $\rm0.3\ s$ & $\rm0.4\ s$ & $\rm0.5\ s$ & $\rm0.6\ s$ & $\rm0.7\ s$ & $\rm0.8\ s$ & $\rm0.9\ s$ & $\rm0.3\ s$  & $\rm0.4\ s$  & $\rm0.5\ s$  & $\rm0.6\ s$  & $\rm0.7\ s$ & $\rm0.8\ s$ & $\rm0.9\ s$ \\ \hline\hline
0.3 s          & 30.9             & 33.0                     & 27.0  & 32.9  & 38.6  & 42.1  & 44.9  & 47.2 & 49.2 & \textbf{26.5}   & 32.7   & 39.0   & 43.5   & 46.1 & 48.3 & 50.3 \\ \hline
0.4 s          & 32.4             & 34.6                     & 34.2  & 28.9  & 32.8  & 36.6  & 39.7  & 41.3 & 43.2 & 34.0   & \textbf{28.5}   & 32.9   & 37.8   & 40.2 & 42.9 & 45.1 \\ \hline
0.5 s          & 34.5             & 36.0                    & 42.7  & 33.7  & 31.1  & 33.5  & 36.5  & 38.3 & 40.2 & 43.0   & 33.9   & \textbf{30.9}   & 34.6   & 37.4 & 39.7 & 41.5 \\ \hline
0.6 s          & 35.9             & 37.8                     & 48.3  & 40.6  & 34.3  & \textbf{33.3}  & 34.4  & 36.7 & 38.4 & 48.4   & 41.5   & 34.3   & 33.7   & 35.4 & 37.8 & 39.4 \\ \hline
0.7 s          & 38.1             & 39.5                     & 52.0  & 45.4  & 39.5  & 35.3  & \textbf{35.2} & 35.7 & 37.2 & 52.4   & 46.2   & 39.8   & 36.3   & 35.4 & 37.3 & 38.3 \\ \hline
0.8 s          & 39.7             & 40.8                     & 54.1  & 49.0  & 43.6  & 39.4  & 37.0 & \textbf{36.6} & 37.2 & 54.5   & 50.0   & 44.2   & 40.5   & 37.4 & 37.2 & 38.2 \\ \hline
0.9 s          & 41.0             & 42.0                     & 55.6  & 51.3  & 46.8  & 42.6  & 39.9 & 38.1 & \textbf{38.0} & 56.0   & 52.9   & 47.3   & 43.5   & 40.2 & 38.6 & 38.3 \\ \hline
\end{tabular}
\end{center}
\end{table*}

We evaluated the performance of the proposed EAM algorithm in terms of the phone error rate (PER) compared to the single DNN background model (SBM) [30]. An SBM whose computational complexity was the same as that of the single network of the EAM and which consists of seven hidden layers, was designed. Further, for fair comparison, we developed an extended SBM (eSBM) with the same computational complexity as the EAM, and the eSBM consisted of 10 hidden layers with 3990 units. Additionally, two background models, EAM and EJAM, were trained using data recorded from $\rm0.3\ s$ to $\rm0.9\ s$, for which a 792-dimensional feature vector consisting of 11 frames of 24-dimensional filterbank features with delta and delta-delta was used in the input layer. In the target layer, 2021 tied states were used, and the learning rate was 0.0001 with a value of 0.9 for momentum. In jointly training the EJAM, a 792-dimensional feature vector was used with three hidden layers, without the softmax output layer. Each of the hidden layers had 2048 units. The target of the feature mapping DNN was a 792-dimensional feature vector of clean speech, and the learning rate and momentum were initially set to 0.0000001 and 0.9, respectively. All DNNs, including the SBM, eSBM, EAM, and EJAM, were initialized using restricted Boltzmann machine (RBM) pre-training followed by fine-tuning using the error back-propagation algorithm (cross entropy for acoustic modeling and MSE for feature mapping). The mini-batch size was set to be 512 for the stochastic gradient algorithm in the error back-propagation algorithm.

\begin{figure}[tbh]
\begin{center}
\includegraphics[width=9.5cm]{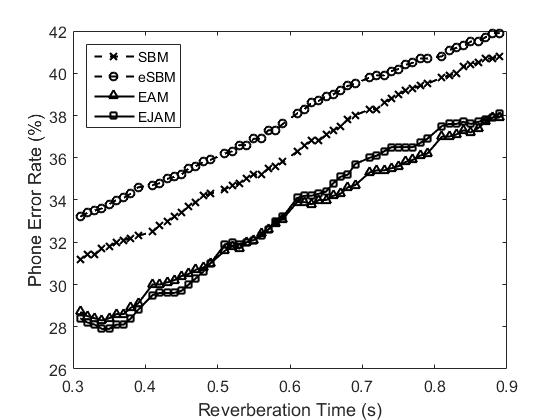}
\caption{Performance [PER (\%)] comparison of the single background models and ensemble models under various reverberant environments}
\end{center}
\end{figure}

In the test step, final acoustic scores were measured for the EAM and EJAM, and were compared to scores of the SBM and eSBM (Figs.\ 3, 5). The proposed ensemble models yielded better performance than the single DNN-based background models in matched test conditions (Table\ I). From this, the single matched model in the proposed ensemble network, including the EAM and EJAM, were superior to the single DNN-based background models for a given $\rm{RT_{60}}$ condition (Fig.\ 7). It is also noteworthy that the eSBM yielded a slightly worse performance than the SBM, particularly for high values of $\rm{RT_{60}}$. This indicates that deeper layers (e.g. eSBM) over the SBM do not exhibit an advantage if the different topological network structures are not presented in high reverberant conditions, unlike our presented ensemble models.

\begin{figure*}[tbh]
\begin{center}
\begin{subfigure}[b]{0.35\textwidth}
\includegraphics[width=\textwidth]{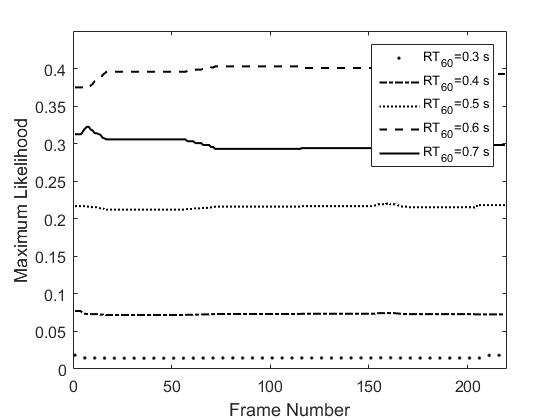}
\caption{\centering ML ($\rm{RT_{60}}=0.61\ s$)}
\end{subfigure}
\begin{subfigure}[b]{0.35\textwidth}
\includegraphics[width=\textwidth]{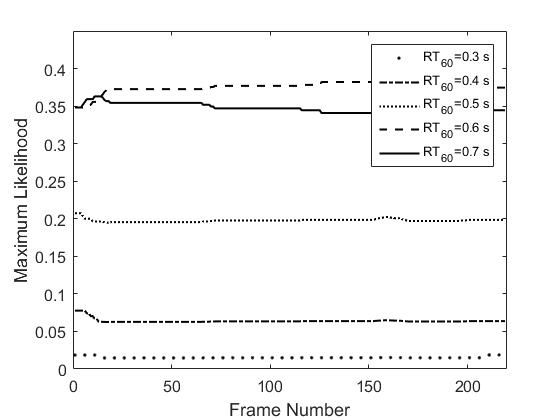}
\caption{\centering ML ($\rm{RT_{60}}=0.63\ s$)}
\end{subfigure}
\begin{subfigure}[b]{0.35\textwidth}
\includegraphics[width=\textwidth]{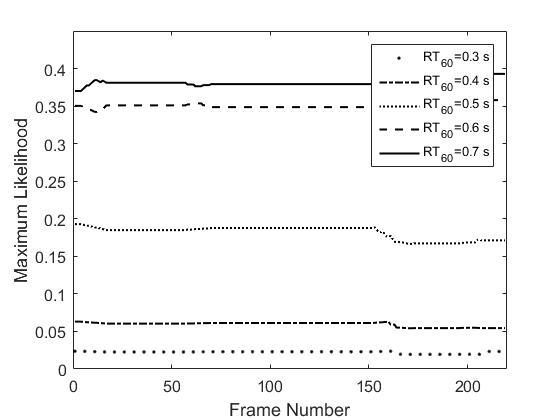}
\caption{\centering ML ($\rm{RT_{60}}=0.67\ s$)}
\end{subfigure}
\caption{An example of the ML of the estimated RT}
\end{center}
\end{figure*}

In addition, we examined the performance of the EAM and EJAM in an online test condition. For this, the $\rm{RT_{60}}$ was blindly estimated at runtime in an ML fashion in order to find the weights for the ensemble structure. This evaluation was performed for $\rm{RT_{60}}$ values between $\rm0.3\ s$ and $\rm0.9\ s$ in increments of $\rm0.01\ s$. ML-based estimates are displayed in Fig.\ 8, and we adopted the parameters used in [29]. The $\rm{RT_{60}}$ estimates of test speech at each $\rm{RT_{60}}$ condition are plotted in Fig.\ 8. The estimated ML of each $\rm{RT_{60}}$ condition did not yield drastic changes; therefore, we used the average of the ML estimates over frames for each utterance. For example, as shown in Fig.\ 8(c), when the utterance was generated in the $\rm{RT_{60}}=0.67\ s$ condition, the two most likely $\rm{RT_{60}}$ conditions ($\rm{RT_{60}}=0.6\ s$ and $\rm{RT_{60}}=0.7\ s$) were selected by two types of largest ML. Further, the ML estimate for $\rm{RT_{60}}=0.7\ s$ was larger than the ML estimate for $\rm{RT_{60}}=0.6\ s$, so the weight of $\rm{RT_{60}}=0.7\ s$ became larger than the weight of $\rm{RT_{60}}=0.6\ s$. Additionally, it appears that two weights computed by ML estimates, as in (8) and (9), are changed over the test utterance. This makes it possible to use this estimator in an online scenario. It is evident from Fig.\ 7 that the EAM achieved relative PER reductions of 8.44\% over the SBM and 12.37\% over the eSBM. It is notable that the eSBM does not yield improved performance when compared to the EAM, though it utilized the same amount of computational complexity. It is clear that the proposed ensemble methods have stronger generalization abilities at runtime conditions, made possible by choosing the best matched models with enhanced sensitivity.

As for the comparison between the EAM and EJAM, it is clear that the EJAM is superior to the EAM when the reverberant time is normal ($\rm{RT_{60}}\le 0.5\ s$). However, when acoustic reverberation is severe ($\rm{RT_{60}}\gneq 0.5\ s$), the EJAM performed slightly worse than the EAM. One possible explanation is that the presented DNN-based feature mapping is not good at representing very long reverberation, as it requires a long-term feature vector as an input of the DNN in order to characterize the long-term evolution of reverberant speech. Future work may benefit from additional features, which might be good at reflecting the long-term pattern of reverberant speech. To summarize, the prepared ensemble models are effective for distant speech recognition in various reverberant conditions. Moreover, performance is improved when using the EJAM in normal reverberant environments.

\section{CONCLUSIONS}
To the best of our knowledge, this study is the first to use an ensemble model for speech recognition under reverberant conditions. The ensemble structure was initially built where each $\rm{RT_{60}}$ in a specific range served as a single DNN-based acoustic modeling. Then, the two most likely DNNs from the ensemble model are chosen based on the ML blind estimation for $\rm{RT_{60}}$. The EAM is further enhanced by jointly training the acoustic and feature models, which serve as the dereverberation. From the experimental results, the DNN-based ensemble structure for acoustic modeling was superior to the single DNN-based background model. In particular, the blind ML estimation for $\rm{RT_{60}}$ was successfully responsible for choosing the most likely DNNs in the ensemble model. The design of the EJAM permitting improved performance in normal reverberant conditions was verified in terms of speech recognition accuracy.

\ifCLASSOPTIONcaptionsoff
  \newpage
\fi

%




\end{document}